\documentclass[10pt,twocolumn,letterpaper]{article}

\usepackage{cvpr}
\usepackage{times}
\usepackage{epsfig}
\usepackage{graphicx}
\usepackage{amsmath}
\usepackage{amssymb}

\usepackage{subfigure}

\usepackage{amsmath}
\usepackage[ruled,longend]{algorithm2e}

\usepackage{times}
\usepackage{epsfig}
\usepackage{graphicx}
\usepackage{amsmath}
\usepackage{amssymb}
\usepackage{verbatim}
\usepackage{subfigure}

\usepackage{times}
\usepackage{courier}
\usepackage{amsmath}
\usepackage{amssymb}
\usepackage{amsfonts}
\usepackage{indentfirst}
\usepackage{xcolor}
\usepackage{multirow}
\usepackage{float}
\usepackage{graphicx}
\usepackage{graphics}
\usepackage{subfigure}
\usepackage{float}
\usepackage{bigstrut}
\usepackage{booktabs}
\usepackage{bibentry}

\usepackage[pagebackref=true,breaklinks=true,letterpaper=true,colorlinks,bookmarks=false]{hyperref}

 \cvprfinalcopy 

\def\cvprPaperID{4606} 

\ifcvprfinal\fi
\begin{document}

\title{Deblurring by Realistic Blurring}

\author{
Kaihao Zhang$^{1}$ \ Wenhan Luo$^2$ \ Yiran Zhong$^{1,4}$ \  Lin Ma$^2$ \ Bjorn Stenger$^{3}$ \ Wei Liu$^2$ \ Hongdong Li$^{1,4}$ \\
$^1$ Australian National University \ \
$^2$ Tencent AI Lab \ \
$^3$ Rakuten Institute of Technology \ \ 
$^4$ ACRV\\ 
}





\maketitle

\pagestyle{empty}  
\thispagestyle{empty} 

\begin{abstract}

Existing deep learning methods for image deblurring typically train models using pairs of sharp images and their blurred counterparts. However, synthetically blurring images do not necessarily model the genuine blurring process in real-world scenarios with sufficient accuracy. To address this problem, we propose a new method which combines two GAN models, \textit{i.e.}, a learning-to-Blur GAN (BGAN) and learning-to-DeBlur GAN (DBGAN), in order to learn a better model for image deblurring by primarily learning how to blur images. The first model, BGAN, learns how to blur sharp images with unpaired sharp and blurry image sets, and then guides the second model, DBGAN, to learn how to correctly deblur such images. In order to reduce the discrepancy between real blur and synthesized blur, a relativistic blur loss is leveraged. As an additional contribution, this paper also introduces a Real-World Blurred Image (RWBI) dataset including diverse blurry images. Our experiments show that the proposed method achieves consistently superior quantitative performance as well as higher perceptual quality on both the newly proposed dataset and the public GOPRO dataset.
\end{abstract}

\begin{figure}[tb]
  \centering
  \subfigure[]{
    \label{idea:blur}
    \includegraphics[width= 0.99\linewidth]{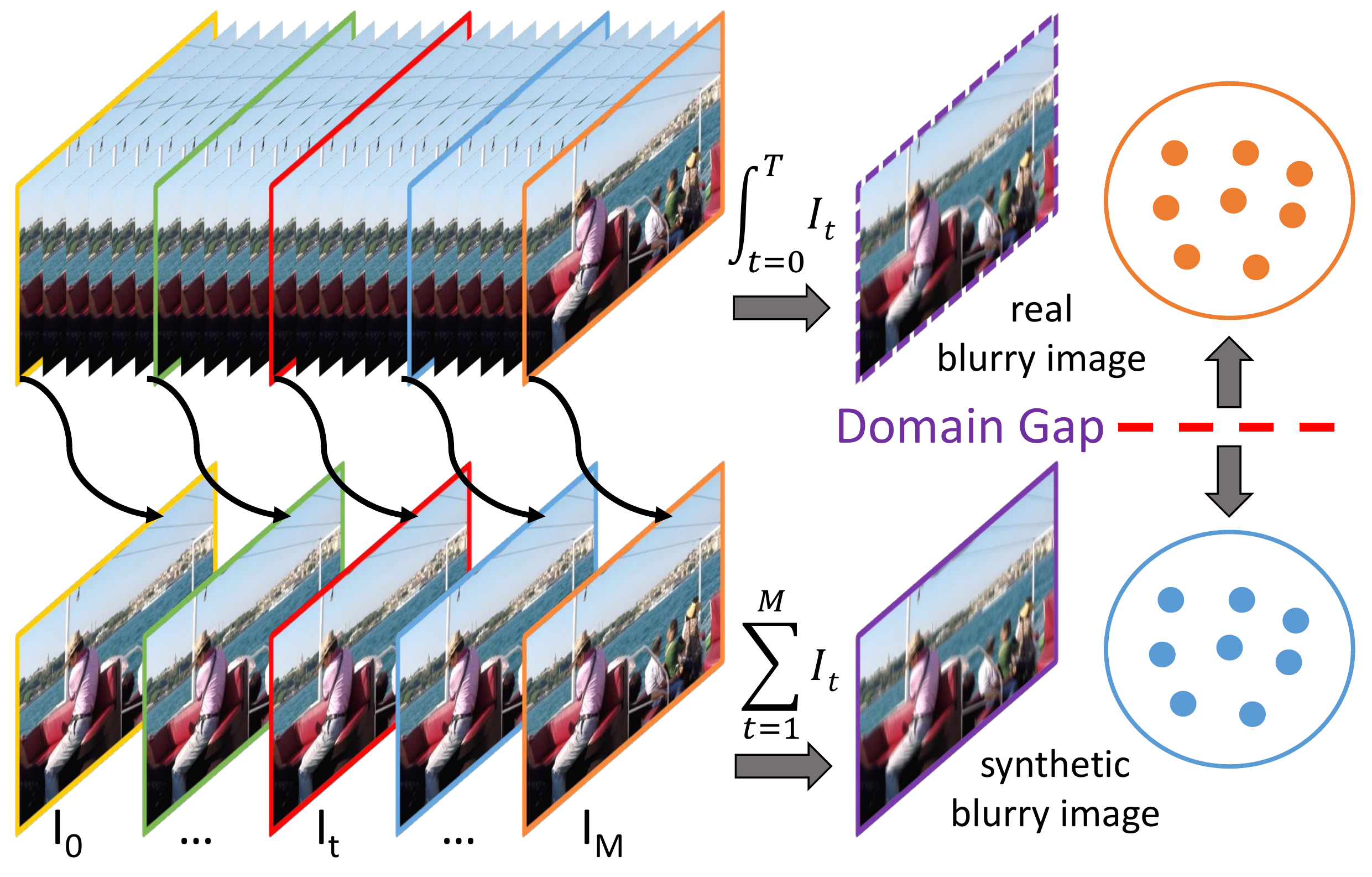}}
  \subfigure[]{
    \label{idea:method}
    \includegraphics[width=0.99\linewidth]{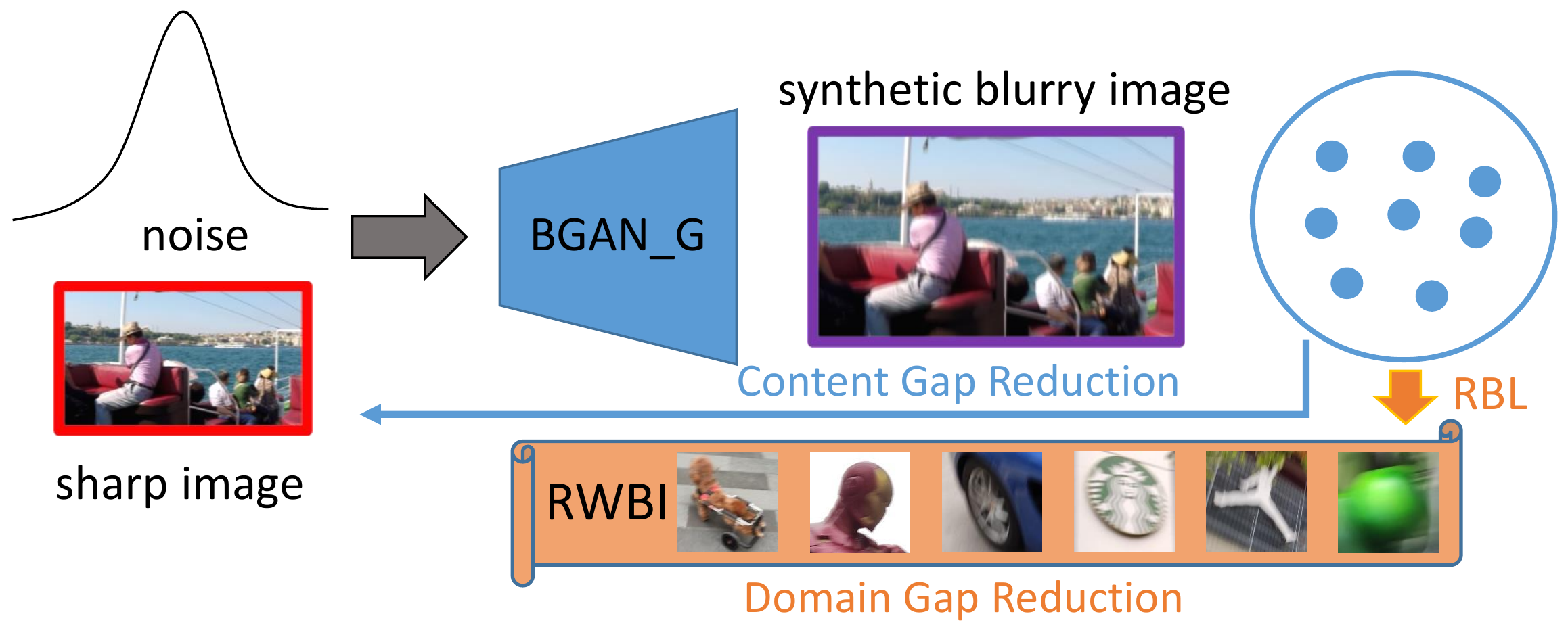}}
  \caption{\small (a) The differences between real and synthetic blurry images; (b) an illustration of learning to blur. Sharp images and random noises are fed into the BGAN\_G model to generate realistic blurry images via the RBL loss and the RWBI dataset.}
  \label{figure_idea}
\end{figure}

\section{Introduction}
\label{introduction}

Image deblurring is a classic problem in low-level computer vision, and it remains an active topic in the vision research community. Given a blurred image, which is corrupted by some unknown blur kernel or a spatially variant kernel, the task of (blind) image deblurring is to recover the sharp version of the original image,  by reducing or removing the undesirable blur in the blurred image.  Traditional deblurring methods handle this problem via estimating a blur kernel, through which a sharp version of the blurred input image can be recovered. Often, special characteristics of the blur kernel are assumed, and natural image priors are exploited in the deblurring process~\cite{cho2009fast,goldstein2012blur,pan2014deblurring,xu2010two,xu2013unnatural}. However, estimating the optimal blur kernel is a difficult task and can therefore impair the overall performance.

Recently, deep learning methods, particularly convolutional neural networks (CNNs), have been applied to tackle this task and obtained a remarkable success, \textit{e.g.}, \cite{nah2017deep,su2017deep,tao2018scale,zhang2018dynamic}. Existing deep learning methods focus on training deblurring models using {\em paired} blurry and sharp images. For example, Nah \textit{et al.} \cite{nah2017deep} propose a multi-scale loss function to implement a coarse-to-fine processing pipeline. Tao \textit{et al.} \cite{tao2018scale} and Gao \textit{et al.} \cite{gao2019dynamic} improve the work by using shared network weights among different scales, achieving state-of-the-art performance.

However, many common effects are not adequately captured by the current deep learning models in the following sense. First, since in real-world scenarios, an image is captured during a time window (\textit{i.e.}, the exposure duration), the blurred image is in fact the integration of multi-frame instant and sharp snapshots~\cite{hirsch2011fast}.  This can be formulated as
\begin{equation}
\label{blurr_process1}
I_B  = g \left(\frac{1}{T}{\int_{t=0}^{T}I_{S(t)}\mathrm{d}t} \right), 
\end{equation}
where $I_S$ is an instant sharp frame and $I_B$ is the blurry image. $T$ is the exposure time period and $g(\cdot)$ is the Camera Response Function (CRF). 
In contrast, in conventional deblurring methods, blurry images used in the training set are often artificially synthesized by approximating the integration step with a simple averaging operation, as shown in Eq. \eqref{blurr_process2}, where $M$ is the number of frames: 
\begin{equation}
\label{blurr_process2}
I_B \simeq g \left( \frac{1}{M}\sum_{t=1}^{M}I_{S[t]} \right).
\end{equation}
Prior methods use $M$ sharp frames $I_{S[t]}$ to replace the continuous sequence $I_{S(t)}$ and generate paired training data, avoiding the complexity of obtaining pairs of real blurry and sharp images. However, there is a clear gap between real blurry images and those artificially blurred images. Fig. \ref{idea:blur} shows the generation of real and synthetic blurry images.

Second, in real situations there are multi-fold factors (not limited to a single linear integration or summation) which can cause image blurs, for instance, camera shake, fast object motion, and small aperture with a wide depth of field. Many of these factors are very difficult to model precisely. To design a better deblurring algorithm, all these factors should be taken into consideration. If the real blurred images are different from the samples in the training set, the trained model may not perform well on the testing data. This observation inspires us to develop a new deblurring method which does not assume any particular blur type; rather such a method will be able to learn a blurring process in order to achieve better deblurring quality.

Specifically, in this paper we propose a method which contains a leaning-to-Blur GAN (BGAN) module and a learning-to-DeBlur GAN (DBGAN) module. BGAN and DBGAN are two complementary processes, in the sense that BGAN learns to mimic properties of real-world blurs by generating photo-realistic blurry images. This module is trained using unpaired sharp and blurry images, thus relaxing the requirement of needing paired data. Recently, Shaham \textit{et al.} propose SinGAN \cite{shaham2019singan} to produce different images based on random noises, which inspires us to generate various blurry images given different noises. During the generation, sharp images are also fed into BGAN to make the generated blurry images bear the same content as the input images. The DBGAN module learns to recover sharp images from blurry images with real sharp and generated blurry images. We further employ a relativistic blur loss, which helps predict the probability that a real blurry image is relatively more realistic than a synthesized one. Finally, a Real-World Blurry Image (RWBI) dataset is created to help train the BGAN model and evaluate the performance of our proposed image deblurring model. Fig. \ref{idea:method} shows the process of learning realistic blur.

The contributions of this paper are three-fold:
(1) We develop a new image deblurring framework which contains the process of image blurring and image deblurring. In contrast to previous deep learning methods which solely focus on image deblurring, our framework also considers image blurring, which generates realistic blurry images to help enhance the performance and robustness of image deblurring.
(2) In order to train the BGAN model and generate blurry images like those in the real world, a relativistic blur loss is introduced. We also contribute a real-world blurry dataset RWBI, which can be used for training an image blurring module and for evaluating deblurring models.
(3) Experimental results show that the proposed method achieves not only the state-of-the-art quantitative performance on the public GOPRO benchmark, but also consistently superior perceptual quality on real-world blurry images.

\begin{figure*}[tb]
  \centering
\includegraphics[width=0.99\linewidth ]{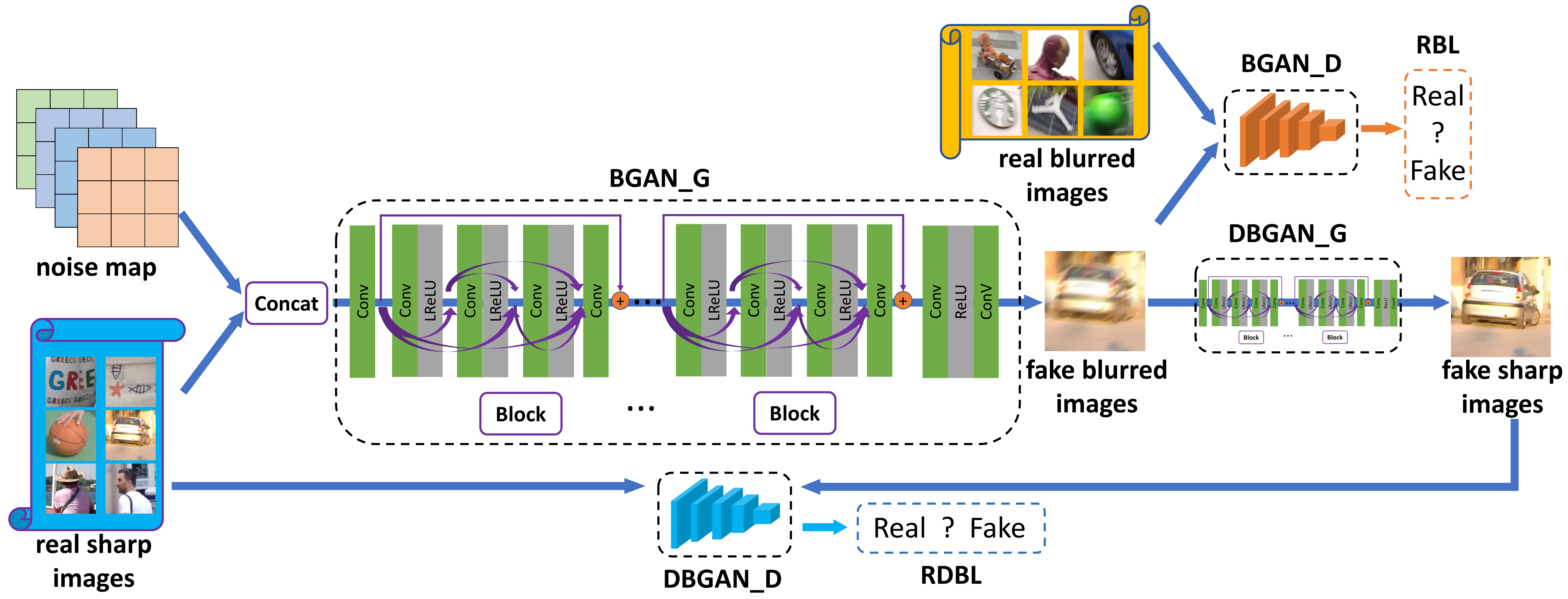}
\caption{{\bf The proposed framework and training process.} This framework contains two main modules, a BGAN and a DBGAN. $D$ and $G$ denote discriminator and generator networks, respectively. The BGAN takes sharp images as input and outputs realistic blurry images, which are then fed into the DBGAN in order to learn to deblur. During the inference stage, only the DBGAN is applied.}
\label{overall}
\end{figure*}

\section{Related Works}
Our work in this paper is closely related to image blurring and image deblurring, which are briefly introduced as follows, respectively.

\subsection{Image Blurring}
Blur artifacts are caused by various factors.  The blurring process can be mathematically formulated as~\cite{gupta2010single,whyte2012non},
\begin{equation}
\label{kernel}
I_B = K \ast I_S + N\,,
\end{equation}
where $I_{B}$ and $I_{S}$ are blurry and sharp images, respectively. $K$ is the unknown (blind) or known (non-blind) blur kernel and $N$ is additive noise. 
For images with spatially varying blurs there are no camera response function (CRF) estimation techniques~\cite{tai2013nonlinear}. Alternatively, the CRF can be approximated as the average of known CRFs as follows:
\begin{equation}
\label{camera}
g(I_{S[i]}) = {I_{S'[i]}}^{\frac{1}{\gamma}}\,,
\end{equation}
where $\gamma$ is a parameter. The latent realistic sharp images $I_{S[i]}$ can be obtained based on the observed sharp images $I_{S'[i]}$. The blurry images can then be generated based on Eq. \eqref{blurr_process2}. Eq. \eqref{blurr_process2} and Eq. \eqref{kernel} are the two main methods to generate image pairs for training. However, neither of them is able to synthesize realistic blurry images like Eq. \eqref{blurr_process1}.

\subsection{Image Deblurring}
Early works use image priors, including total variation~\cite{chan1998total}, a heavy-tailed gradient prior~\cite{shan2008high}, or a hyper-Laplacian prior~\cite{krishnan2009fast}, which are typically applied to images in a coarse-to-fine manner. Recently, deep learning methods have achieved a great success in the areas of object recognition \cite{he2016deep,zhang2019learning,zhang2019cousin,li2020transferring,li2020word} and image reconstruction including video deblurring \cite{zhang2018adversarial}, video dehazing~\cite{ren2018deep}, and other GAN-based generation tasks ~\cite{shen2018deep,ren2018gated,xiong2019foreground,wang2018esrgan}. For image deblurring, Sun \textit{et al.}~\cite{sun2015learning} propose a CNN-based model to estimate a kernel and remove non-uniform motion blur. Chakrabarti~\cite{chakrabarti2016neural} uses a network to compute estimations of sharp images that are blurred by an unknown motion kernel. Nah \textit{et al.}~\cite{nah2017deep} propose a multi-scale loss function to apply a coarse-to-fine strategy and an adversarial loss. Kupyn \textit{et al.} propose DeblurGAN \cite{kupyn2018deblurgan} and DeblurGAN-v2 \cite{kupyn2019deblurgan} to remove blur kernels based on adversarial learning. Further, RNN-based methods have been proposed for image deblurring. Zhang \textit{et al.}~\cite{zhang2018dynamic} propose a spatially variant neural network, which includes three CNNs and one RNN. Tao \textit{et al.}~\cite{tao2018scale} propose an SRN-DeblurNet, which includes one LSTM and CNNs for multi-scale image deblurring. 
Shen \textit{et al.} \cite{shen2019human} introduce a human-aware deblurring method to remove blur from foreground humans and background. Gao \textit{et al.}~\cite{gao2019dynamic} propose a nested skip connection structure which achieves state-of-the-art performance.

 All these above neural network based methods focus on solely recovering sharp images from blurry images (\textit{i.e.}, image deblurring), rather than better modeling the blurring process itself. Pan \textit{et al.} \cite{pan2018physics} try to generate blurry images in their algorithm based on deblurred results, and then calculate the difference between the generated blurry images and ``GT" blurry images to update models. Therefore, these methods actually propose a new loss function, rather than data augmentation. The idea of data augmentation has been widely applied in different fields \cite{shorten2019survey}, like face verification \cite{liu2018exploring} and SR \cite{bulat2018learn}. For deblurring, one of the most relevant works is from the field of video deblurring \cite{chen2018reblur2deblur}. However, it generates blurred images based videos and it does not consider to generate realistic blurry images based on real blurred ones. More recently, a SinGAN \cite{shaham2019singan} model is proposed to learn how to generate different related images from one input image based on random noises. Inspired by this method, a GAN-based model is proposed to generate various blurry images based on different noises.

\section{Deblurring by Blurring}
\label{sec:approach}

\subsection{Overall Architecture}
\label{sec:architecture}

Our framework contains two primary modules. Similar to prior image deblurring works, our framework includes a learning-to-DeBlur GAN (DBGAN) module, which is trained on paired sharp and blurry images to recover sharp images from blurry images. The paired sharp-blurry images are obtained from the BGAN module. The BGAN is trained on unpaired data, where sharp images come from a public dataset, while the blurry images come from a new real-world blurry dataset. Fig. \ref{overall} shows the overall architecture of the proposed framework.

We further enhance the standard GAN model with a relativistic blur loss. In traditional GAN-based models for image deblurring, the discriminator $D$ estimates the probability that the input data is real, and the generator $G$ is trained to increase the probability that the generated data looks real. The developed relativistic blur loss estimates the probability that the given real-world blurry images are more realistic than the generated blurry images.

In the training stage, sharp images are input into the BGAN generator and its output is fed into the DBGAN to learn how to deblur. The generators in the DBGAN and BGAN modules generate corresponding images, and the discriminators conduct discrimination to create more realistic synthetic images. During the inference stage, only the DBGAN generator network is required for the image deblurring task.

\subsection{BGAN: Learning to Blur}
\label{BGAN}
The BGAN module is the primary difference from other neural network based methods for image deblurring. Similar to other GAN based models, the BGAN consists of a generator network and a discriminator network. In this section, we first discuss its architecture and loss functions.

\textbf{BGAN Generator}.
The input to the BGAN generator is a sharp image from a public dataset. 
Given the numerous possible factors that can cause undesired blurring artifacts, we concatenate the input image with a noise map to model the different conditions. To obtain the noise map, we sample a noise vector of length $4$ from a normal distribution and duplicate it $128 \times 128$ times in the spatial dimension to obtain a $4 \times 128 \times 128$ noise map as in~\cite{zhu2017toward}. In this way, we can generate various blurry images based on one sharp image. The network architecture consists of one convolutional layer, $9$ residual blocks (ResBlocks)~\cite{he2016deep} and extra two convolutional layers. Each ResBlock consists of $5$ convolutional layers ($64 \times 3 \times 3$) and $4$ ReLU activations. There is also a skip connection in each ResBlock, connecting the input and output features (refer to Fig. \ref{overall}). The output of our BGAN generator is a blurry image of the same size as the sharp input image.

\begin{figure*}[t]
  \centering
\includegraphics[width=0.83\linewidth ]{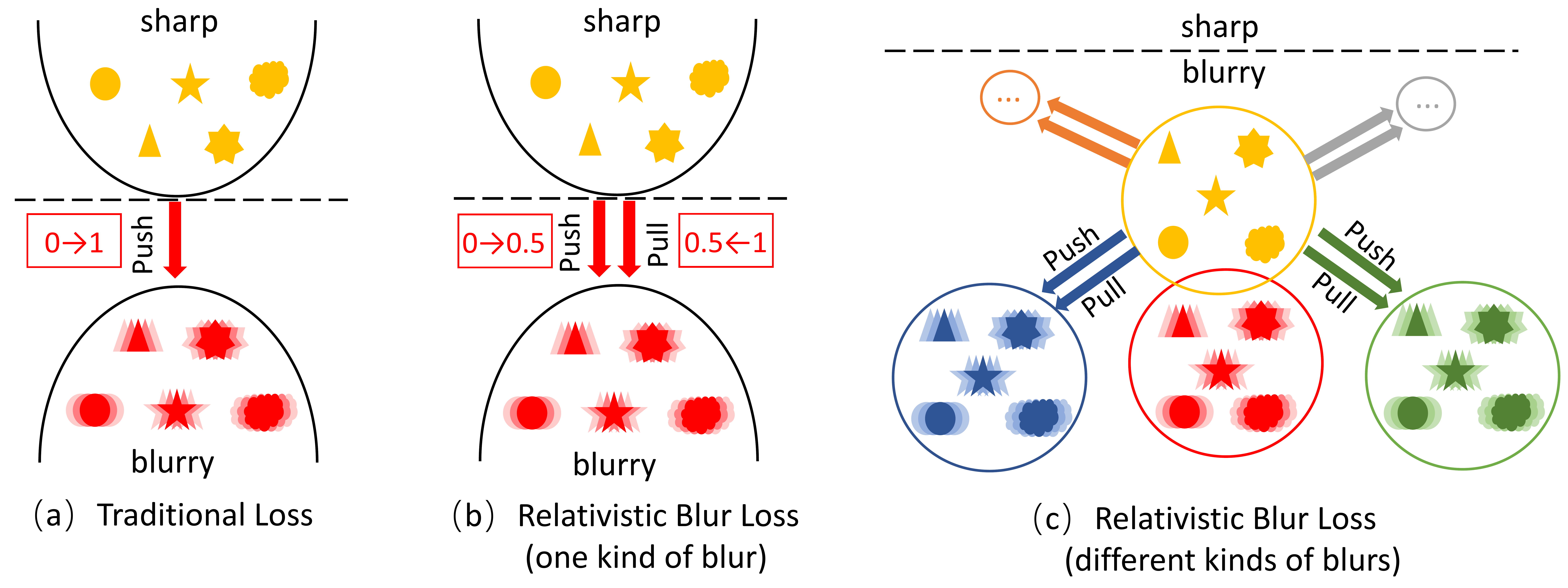}
\caption{{\bf An illustration of the Relativistic Blur Loss (RBL).}
Real and synthesized images are labeled as 1 and 0, respectively. (a) A traditional loss function is used to update the generator to create blurry images (label=0) which are similar to real ones (label=1). (b) The RBL not only increases the probability that generated images look real (0 $\rightarrow$ 0.5, which is labeled as ``Push"), but also simultaneously decreases the output probability that real images are real (1 $\rightarrow$ 0.5, which is labeled as ``Pull"). 
(c) In order to increase the variations of blurry images, different blurry images are used to model the different types of blurs in the real world.}
\label{RBL}
\end{figure*}

\textbf{BGAN Discriminator}. The input to the BGAN discriminator is the output of the BGAN generator. Its architecture is the same as the VGG19 network~\cite{simonyan2014very}, and its output is the probability of the blurry image being classified as real.

\textbf{BGAN Loss}. The generator and discriminator of the BGAN are trained with a perceptual loss and an adversarial loss. Specifically, the perceptual loss is calculated based on the synthesized blurry images from the proposed BGAN and images taken from a public dataset. In this way, they can have similar contents. The adversarial loss is calculated between the synthesized and real blurry images. The real blurry images are taken from our newly created dataset.

\subsection{DBGAN: Learning to Deblur}
\label{sec:DBGAN}
The BGAN module aims to mimic the real-world blurry images and cover as many blur cases as possible. Its goal is to drive the DBGAN module to be more effective in recovering sharp images from blurry images. In the following, we present the architecture and loss of the DBGAN.

\textbf{DBGAN Generator}. The input to the DBGAN generator is a blurry image.  Many approaches have been proposed for this task~\cite{chakrabarti2016neural,nah2017deep,sun2015learning,tao2018scale}. When we design the DBGAN generator, we adopt their advantages.
Specifically, we remove the batch normalization layers, which have been shown to increase the computational complexity and decrease the performance on different tasks~\cite{nah2017deep}. Secondly, we use additive residual layers in each block, which combine multi-level residual networks and dense connections ~\cite{huang2017densely}. The BGAN consists of one convolutional layer, $16$ residual blocks (ResBlocks)~\cite{he2016deep} and two more convolutional layers. The kernel size in ResBlocks is $63 \times 3 \times 3$. The details can be referred to Fig. \ref{overall}. The output of the DBGAN generator is the desired sharp image.

\textbf{DBGAN Discriminator}. Similar to the BGAN discriminator, the DBGAN also adopts the VGG19 network~\cite{simonyan2014very} as its discriminator. The output of this model is the probability of  the given sharp images looking realistic.

\textbf{DBGAN Loss}. Like the BGAN module, the proposed DBGAN model is trained using a perceptual loss and an adversarial loss. We also use an $L_1$ loss to update the DBGAN. All the three types of loss functions are calculated based on the generated and real sharp images, so the DBGAN is trained on paired images.

\subsection{Relativistic Blur Loss}
\label{relativistic}

In this section, we describe a Relativistic Blur Loss (RBL) and other loss functions which are used to train our framework.

\textbf{Perceptual Loss}. In contrast to previous image deblurring methods~\cite{nah2017deep,tao2018scale}, the proposed framework applies a perceptual loss $\mathcal{L}_{perceptual}$ to update models. Note that Johnson \textit{et al.}~\cite{johnson2016perceptual} use a similar loss. However, in contrast to their work, we calculate the perceptual loss based on features before rather than after the ReLU activation layer.

\textbf{Content Loss}. The Mean Squared Error (MSE) is widely used as a loss function for image restoration methods. Based on the MSE, the content loss between ground-truth and generated images is calculated.

\textbf{Relativistic Blur Loss}. In order to drive the BGAN generator to produce blurry images similar to the real-world images, we develop a relativistic blur loss based on \cite{jolicoeur2018relativistic} to update the model. The BGAN generator parameters are updated in order to fool the BGAN discriminator. The adversarial loss $D$ is formulated as:
\begin{equation}
\centering
\label{RBL_goal}
\begin{array}{l}
\ \ \ \ \ \ \ \ \ \ \ \ \ \ \ \ D({I^{real}_{blurry}}) = \sigma (C({I^{real}_{blurry}})) \rightarrow 1 ,
\\ 
\\
D({I^{fake}_{blurry}}) = D(G({I^{real}_{sharp}})) = \sigma (C(G({I^{real}_{sharp}})))  \rightarrow 0 \,,
\end{array}
\end{equation}
where $D(\cdot)$ is the probability that the input is a real image. 
$C(\cdot)$ is the feature representation before activation and $\sigma(\cdot)$ is the sigmoid function. The generator $G$ is trained to increase the probability that synthesized images are real. Real and synthesized images are labeled as 0 or 1 by $D$, respectively. As Fig. \ref{RBL} (a) shows, the effect of $G$ is to transfer real sharp images to blurry images and "push" these generated images (label=0) closer to real blurry images (label=1). However, during the training stage, only the second part of Eq. \eqref{RBL_goal}, \ie, $D({I^{fake}_{blurry}}) = D(G({I^{real}_{sharp}}))  \rightarrow 0\ $,  updates the parameters of generator $G$, while the first part is used to update the discriminator $D$ model rather than the generator $G$~\cite{nah2017deep}. In fact, a powerful generator $G$ should also decrease the probability that real blurry images are real. This is because a realistic synthesized image labeled as fake is similar to real one, and will thus fool the $D$ model to learn to distinguish real and fake in the training stage. Based on this idea, we add $D({I^{real}_{blurry}})$ into the process of learning $G$ in BGAN. Specially, a Relativistic Blur Loss (RBL) is developed to help calculate whether a real blurry image is more realistic than the synthesized blurry image. The formulation of Eq. \eqref{RBL_goal} is modified to
\begin{equation}
\label{RBL_goal_2}
\begin{array}{l}
\sigma (C({I^{real}_{blurry}}) - E(C(G({I^{input}})))) \rightarrow 1\, , 
\\ 
\\
\sigma (C({I^{fake}_{blurry}}) - E(C({I^{real}_{blurry}}))) \rightarrow 0\,,
\end{array}
\end{equation} 
where $E(\cdot)$ denotes the averaging operation over images in one batch. Fig. \ref{RBL} (b) shows the aim of RBL. Although the goal is still to generate realistic blurry images which are similar to real-world ones, the optimization objective is different. RBL aims to update $G$ to generate synthetic images which are near $0.5$, and meanwhile to fool the $D$ model, making it difficult to distinguish real images from fake ones. In this way, the probability of real blurry images predicted by $D$ is also near to $0.5$. We term the effects as "push" and "pull", respectively, which can complement each other to update the generator $G$. As Fig. \ref{RBL} shows, the sharp and blurry images can be regarded as two different domains. In order to rapidly generate blurry images and utilize prior research results of generating blurry images, we first train our BGAN model with artificially blurry images as Fig. \ref{RBL}(b) shows. We then add other types of blurry images to increase the variations of the produced blurry images based on Eq. \eqref{RBL_goal_2} to cover different conditions in the real world, which is shown in Fig. \ref{RBL}(c).

Based on Eq. \eqref{RBL_goal_2} and Fig. \ref{RBL}, our RBL, which is used in the BGAN generator, can be represented as
\begin{equation}
\small
\begin{array}{l}
{\mathcal{L}}_{RBL} = -[\log (\sigma (C({I^{real}_{blurry}})-E(C(G(I^{input})))) )
\\
\\
\ \ \ \ \ \ \ \ \ \ \ \ \ \ \ \ + \log (1 - (\sigma (C({G(I^{input})})-E(C(I^{real}_{blurry}))))] . \
\end{array}
\end{equation}

Based on the RBL, we apply a Relativistic Deblur loss (RDBL) in the DBGAN generator as
\begin{equation}
\small
\begin{array}{l}
{\mathcal{L}}_{RDBL} = -[\log( \sigma(C({I^{real}_{sharp}})-E(C(G(I^{input}))))) 
\\
\\
\ \ \ \ \ \ \ \ \ \ \ \ \ \ \ \ \ \ + \log (1 - (\sigma({G(I^{input})})-E(C(I^{real}_{sharp})))))] . \
\end{array}
\end{equation}

\textbf{Balance of Different Loss Functions.} During the training stage, the loss functions for DBGAN and BGAN are combinations of different terms using a weighted fusion,
\begin{equation}
\label{L_BGAN}
\mathcal{L}_{BGAN} = \mathcal{L}_{perceptual} + \beta \cdot \mathcal{L}_{RBL} ,
\end{equation}
\begin{equation}
\label{L_DBGAN}
\mathcal{L}_{DBGAN} = \mathcal{L}_{perceptual} + \alpha \cdot \mathcal{L}_{content} + \beta \cdot \mathcal{L}_{RDBL}\,.
\end{equation}
In order to balance the different kinds of losses, we use two hyper-parameters $\alpha$ and $\beta$ to yield the final loss $\mathcal{L}$ for BGAN and DBGAN.

\section{Experiments}

\subsection{Datasets}
\label{dataset}

\textbf{GOPRO Dataset.} We evaluate the performance of our model on the public GOPRO dataset~\cite{nah2017deep}, which contains $3,214$ image pairs. The training and testing sets include $2,103$ and $1,111$ pairs, respectively. Existing methods convolve sharp images with a blur kernel~\cite{chakrabarti2016neural,schuler2016learning,sun2015learning} to synthesized blurry images. These synthetic blurry images are different from real ones captured by camera. In order to model more realistic blurry conditions, in the GOPRO dataset,  sharp images with a high-speed camera and synthesize blurry images were collected by averaging these sharp images from videos. 

\textbf{RWBI Dataset.} 
In order to train our BGAN model and evaluate the performance of deblurring models, we collect a Real-World Blurry Image dataset. The blurry images are captured with different hand-held devices, including an iPhone XS, a Samsung S9 Plus, a Huawei P30 Pro and a GoPro Hero 5 Black. Multiple devices are used to reduce the bias towards one specific device which may capture blurry images with unique characteristics. The dataset contains $22$ different sequences of $3,112$ diverse blurry images. 

We compare the performance of the proposed method with the state-of-the-art methods on the public GOPRO dataset quantitatively and qualitatively. As there is no ground truth of the developed RWBI dataset, we only conduct a qualitative comparison.

\subsection{Implementation Details}
\label{implementation}

When training BGAN and DBGAN, we use a Gaussian distribution with zero mean and a standard deviation of 0.01 to initialize the weights. In each iteration, we update all the weights after learning a mini-batch of size $4$. To augment the training set, we crop a $128 \times 128$ patch at any location of an image. To further increase the number of training samples, we also randomly flip frames. We use a learning rate annealing scheme, starting with a value of $10^{-4}$ and reducing it to $10^{-6}$ after the training loss gets converged. The hyper-parameters $\alpha$ and $\beta$ are set as $0.005$ and $0.01$, respectively.

\begin{figure}[t]
  \centering
\includegraphics[width=0.99\linewidth]{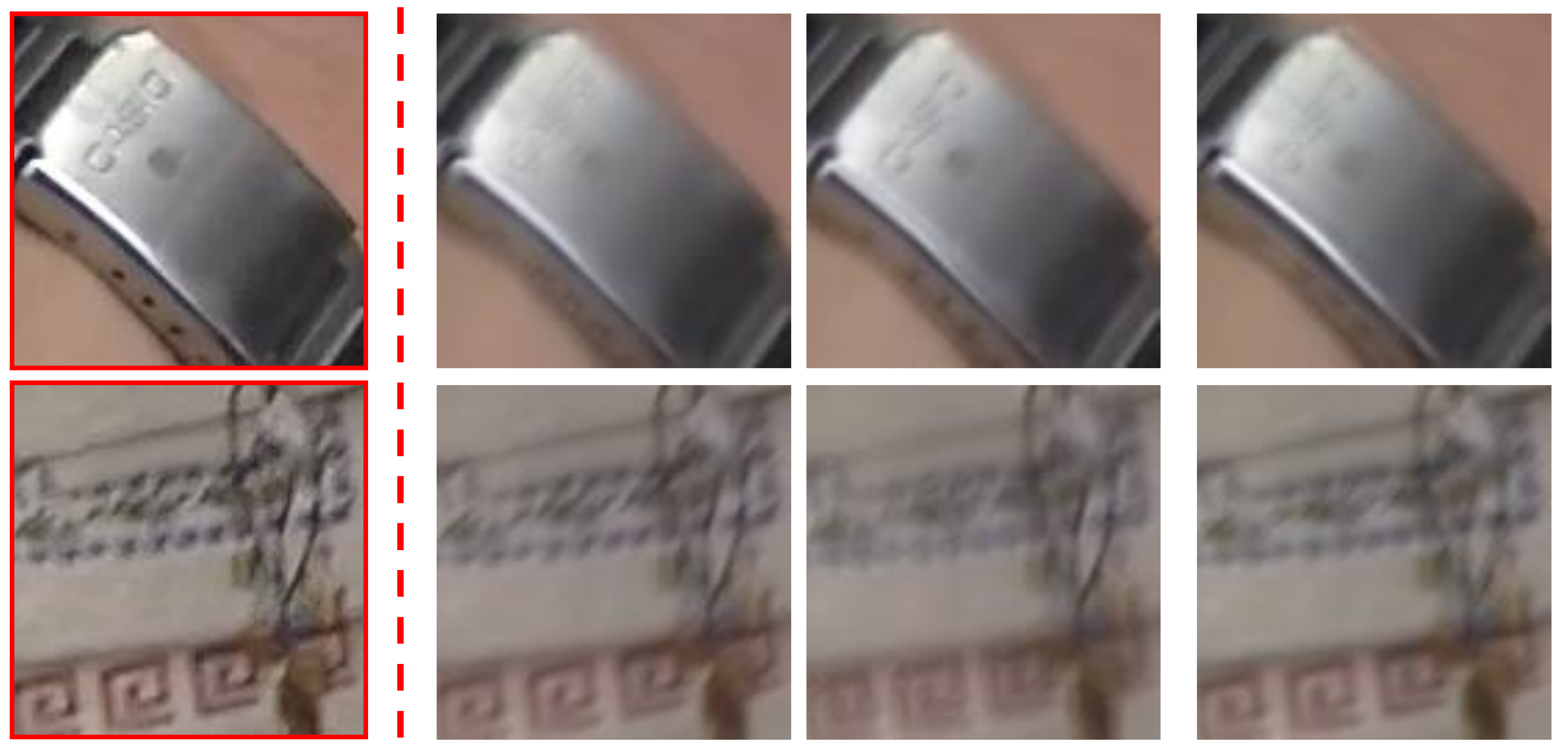}
  \caption{{\bf Synthesized blurry images.} Examples of different blurry images created by the proposed BGAN. The first column shows input sharp images, and the next three columns are the produced blurred images used to train the DBGAN(+).}
  \label{fake}
\end{figure}

\begin{figure}[tb]
  \centering
\includegraphics[width=1\linewidth]{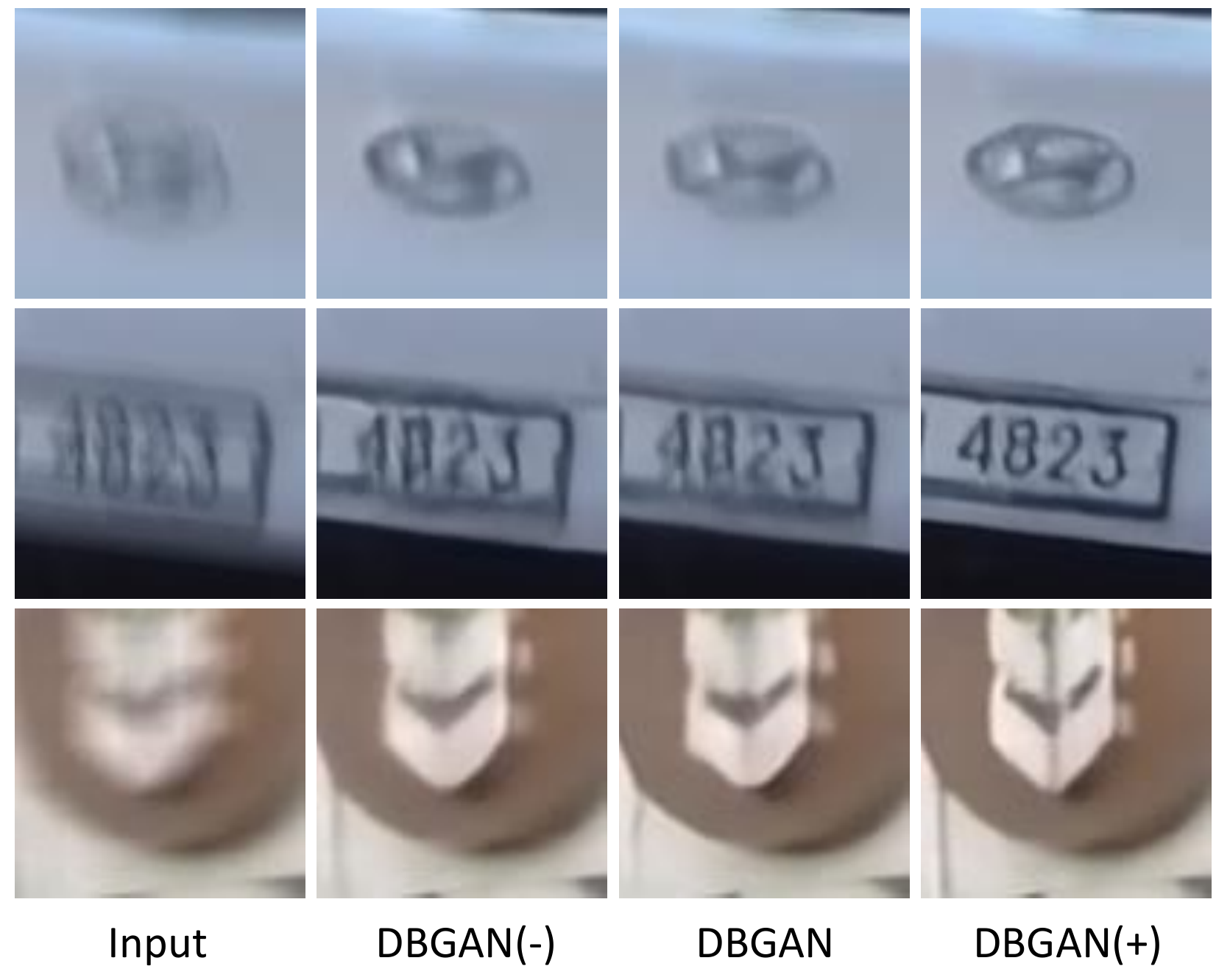}
  \caption{{\bf Qualitative ablation results.} Examples of deblurred images generated by the proposed framework with different model structures. The first column shows input blurred images, and the next three columns are the deblurred images produced by DBGAN(-), DBGAN and DBGAN(+), respectively.}
  \label{figure_ablation}
\end{figure}

\begin{figure*}[tb]
  \centering
\includegraphics[width=1\linewidth]{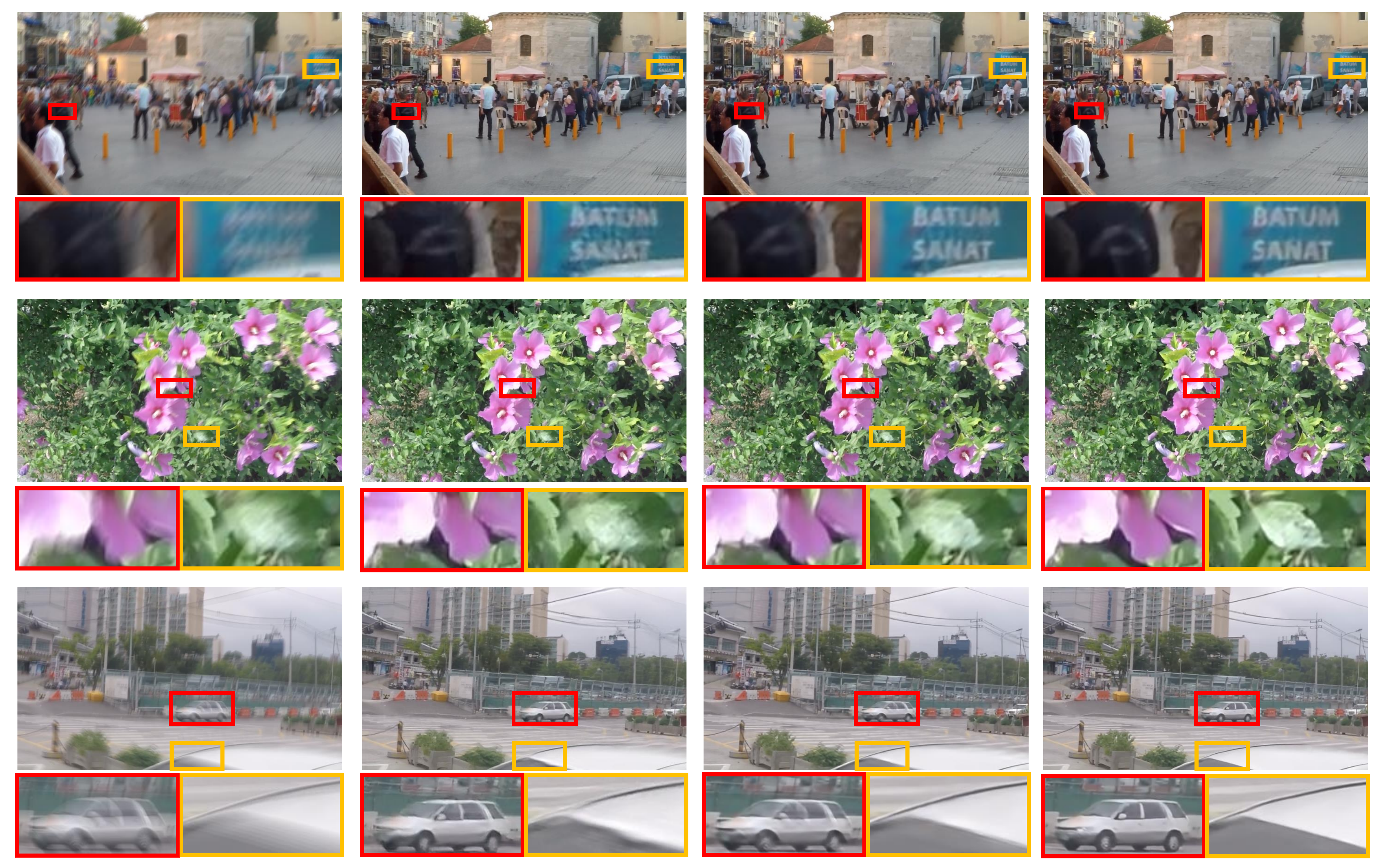}
  \caption{{\bf Comparison with state-of-the-art deblurring methods.} From left to right: blurry images, results of Nah \emph{et al.}~\cite{nah2017deep}, Tao \emph{et al.}~\cite{tao2018scale} and the proposed DBGAN(+) method. The improvement is clearly visible in the magnified patches.}
  \label{qualitative_compare}
\end{figure*}

\subsection{Ablation Study}
\label{comparison}

In this section, we conduct experiments to investigate the effectiveness of different components of our model. The proposed model has three variants:

(1) \textbf{DBGAN} is the model for learning to deblur. Its input is a blurry image and the output is a deblurred image. Similar to previous GAN-based deblurring methods \cite{nah2017deep,kupyn2018deblurgan}, this model contains generator and discriminator networks. Thus its loss function is a combination of ${\mathcal{L}_{percetpual}}$, ${\mathcal{L}_{content}}$ and ${\mathcal{L}_{RDBL}}$ with weights $\alpha$ and $\beta$. The final loss function is shown in Eq. \eqref{L_DBGAN}.

(2) \textbf{DBGAN(-)} has the same architecture as DBGAN. Differently, we replace the ${\mathcal{L}_{RDBL}}$ with a traditional adversarial loss as \cite{nah2017deep}. Namely, the training process does not contain the relativistic loss functions. It is trained based on ${\mathcal{L}_{percetpual}}$, ${\mathcal{L}_{content}}$ and the traditional adversarial loss.  

(3) \textbf{DBGAN(+)} is our full method. It has a similar architecture to DBGAN with the main difference of additionally employing the BGAN module during the fine-tuning stage. Specially, we firstly train a DBGAN model as above, and then blurry images generated by the BGAN model are randomly added into the training samples to enhance the learning performance of DBGAN. Fig. \ref{fake} shows the examples of different blurry images produced by the proposed BGAN.

Fig. \ref{figure_ablation} shows results of the qualitative comparison. The proposed DBGAN outperforms the DBGAN(-), which shows the effectiveness of the relativistic loss function for image deblurring. With the learning-to-blur module, DBGAN(+) achieves a further improvement over DBGAN, suggesting the benefits of learning to deblur by learning to blur.

\subsection{Comparison with Existing Methods}
\label{comparison}

\begin{figure}[]
  \centering
  \subfigure[The blurry image]{
    \label{fig_gao:a}
    \includegraphics[width=1.02in]{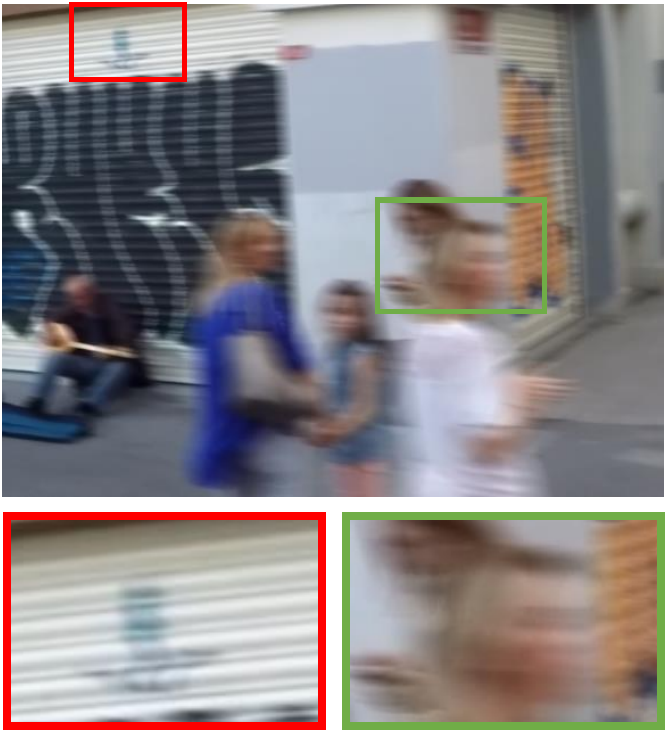}}
  \subfigure[Gao \textit{et al.}$^*$ \cite{gao2019dynamic}]{
    \label{fig_gao:b}
    \includegraphics[width=1.02in]{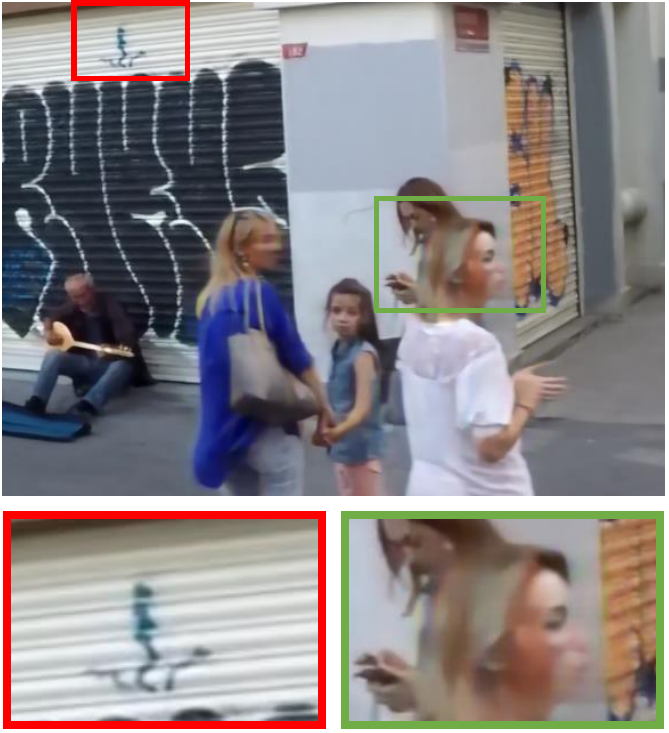}}
  \subfigure[Ours]{
    \label{fig_gao:c}
    \includegraphics[width=1.02in]{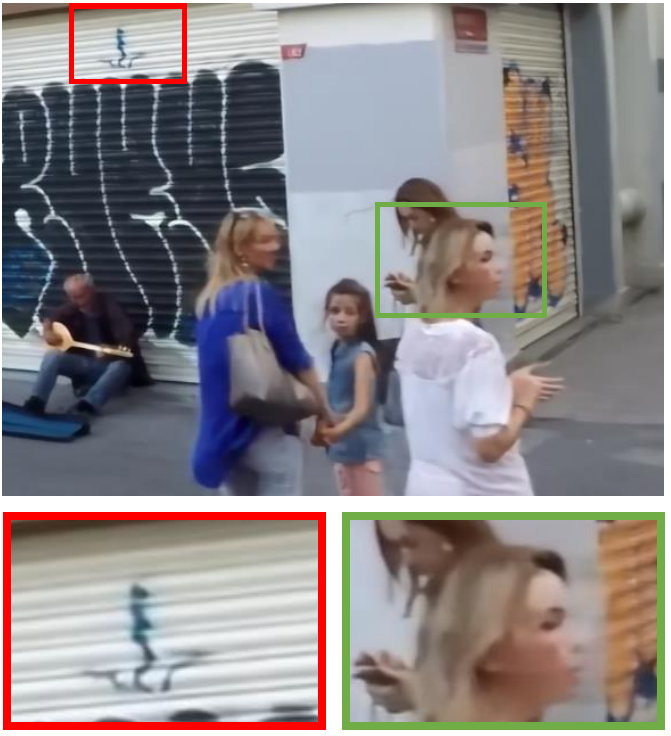}}
  \caption{ \bf Comparison with \cite{gao2019dynamic}, which is trained with extra pairs of images.}
  \label{fig_gao}
\end{figure}

To verify the effectiveness of our model, we compare its performance with several state-of-the-art approaches on the GOPRO dataset quantitatively and qualitatively. \cite{hyun2013dynamic} by Kim \textit{et al.} is a traditional method to handle complex dynamic blurring images. For deep learning methods, Sun \textit{et al.}~\cite{sun2015learning}  use a CNN network to estimate blur kernels and apply traditional deconvolution methods to synthesize sharp images. Nah \textit{et al.}~\cite{nah2017deep} propose a multi-scale function to model the coarse-to-fine approach. Similar to~\cite{nah2017deep}, Tao \textit{et al.}~\cite{tao2018scale} propose a multi-scale network via sharing network weights between different scales to recover sharp images. 
In addition, Shen \textit{et al.} \cite{shen2019human} introduce a human-aware deblurring method and Gao \textit{et al.} \cite{gao2019dynamic} propose a nested skip connection structure and achieve state-of-the-art performance. 
Table \ref{table_compare1} shows the results of the quantitative comparison. DBGAN outperforms most of previous methods, while DBGAN(+) achieves the state-of-the-art performance due to the framework of learning to deblur by learning to blur. For fair comparison, all values refer to the performance achieved by single model trained on the GOPRO dataset. Qualitative comparisons with some state-of-the-art methods are shown in Fig. \ref{qualitative_compare}, demonstrating that our method consistently achieves better visual quality results. Fig. \ref{fig_gao} compare the proposed method with Gao \textit{et al.}$^*$ \cite{gao2019dynamic}. $*$ means this model is trained with extra pairs of images.

\begin{table*}[tb]
  \centering 
    \caption{\it Performance comparison on the {\em GOPRO\_Large} dataset.}
    \setlength\tabcolsep{5.0pt}
    \begin{tabular}{l c c c c c c c c c }
    \toprule
    Method &  Kim \textit{et al.}  &  Sun \textit{et al.} &  Nah \textit{et al.} &  Tao \textit{et al.} & Shen \textit{et al.} & Gao \textit{et al.} & \textbf{DBGAN} & \textbf{DBGAN(+)} \\
    \midrule
    PSNR & 23.64 & 24.64 & 29.08 & 30.10 & 30.26 & 30.92& \textbf{30.43} & \textbf{31.10} \\
    SSIM & 0.8239 & 0.8429 & 0.9135 & 0.9323 & 0.940 & 0.9421& \textbf{0.9372} & \textbf{0.9424}\\
    \bottomrule
    \end{tabular}%
    \label{table_compare1}
\end{table*}%

\begin{figure*}[tb]
  \centering
\includegraphics[width=1\linewidth]{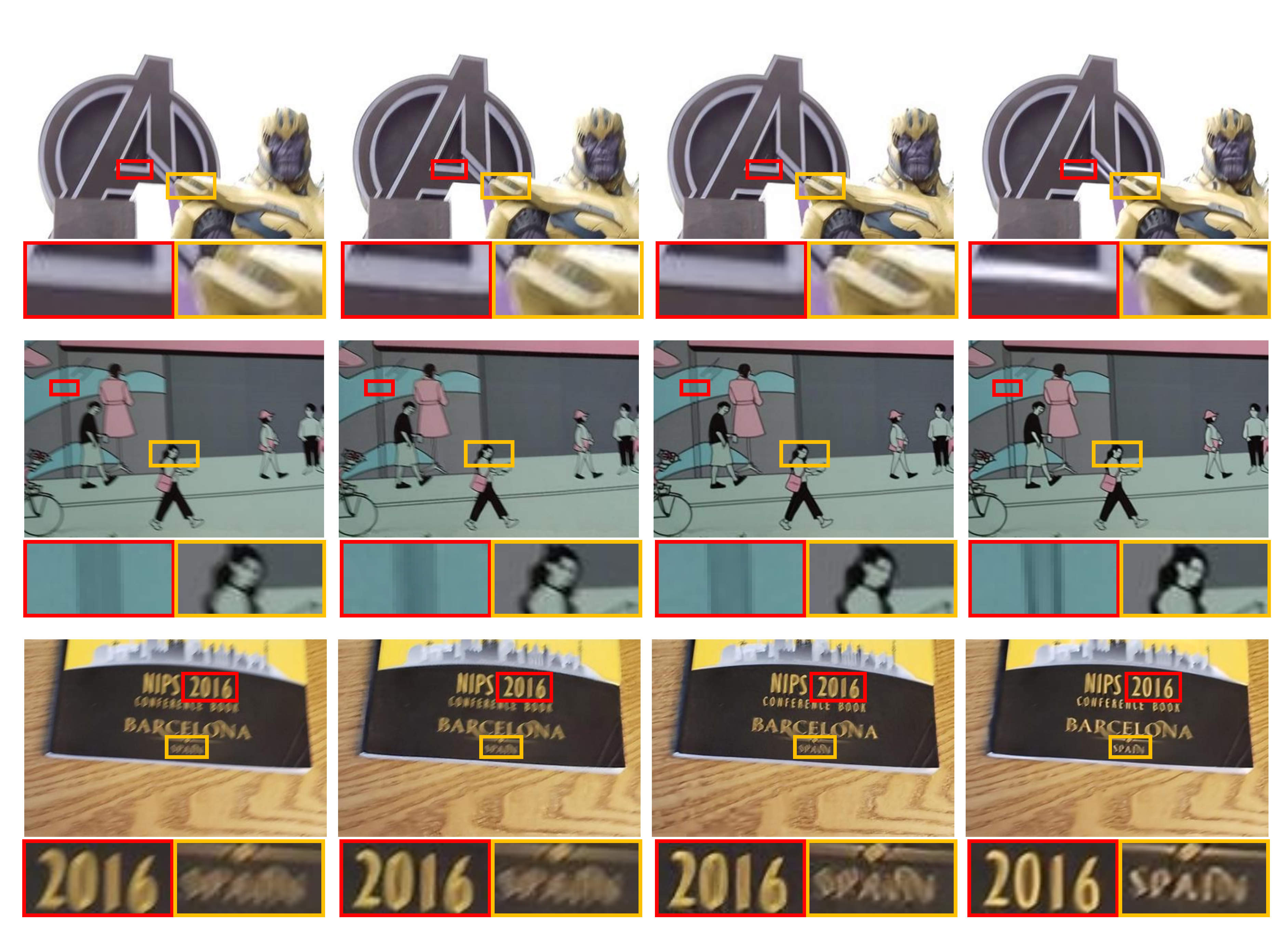}
  \caption{{\bf Performance comparison on real-world blurry images.} From left to right: blurry images, results of Nah \emph{et al.}~\cite{nah2017deep}, Tao \emph{et al.}~\cite{tao2018scale} and the proposed DBGAN(+) method. The improvement is clearly visible in the magnified patches.}
  \label{real_world_fig}
\end{figure*}

\subsection{Performance in Real-World Scenarios}
\label{real_world}

To validate the effectiveness of our method, we compare the performance of our approach with several state-of-the-art methods on the RWBI dataset of real-world blurry images. Fig.~\ref{real_world_fig} shows qualitative results of different models. The blurry images in the first column are from the RWBI dataset, and the images in the following columns are the results of Nah \emph{et al.}~\cite{nah2017deep}, Tao \emph{et al.}~\cite{tao2018scale} and the proposed DBGAN(+). Fig. \ref{real_world_fig} shows that our method achieves better performance on real-world blurry images.

\section{Conclusion}
This paper has presented a new framework which firstly learns how to transfer sharp images to realistic blurry images via a learning-to-blur GAN (BGAN) module. This framework trains a learning-to-deblur GAN (DBGAN) module to learn how to recover a sharp image from a blurry image. In contrast to prior work which solely focuses on learning to deblur, our method learns to realistically synthesize blurring effects using unpaired sharp and blurry images.
In order to generate more realistic blurred images, a relativistic blur loss is employed to help the BGAN module reduce the gap between synthesized blur and real blur. In addition, a RWBI dataset is built to help train and test deblurring models. The Experimental results have demonstrated that our method not only yields results of consistently superior perceptual quality, but also outperforms state-of-the-art methods quantitatively.

\section*{Acknowledgment}
This work is funded in part by the ARC Centre of Excellence for Robotics Vision (CE140100016),  ARC-Discovery (DP 190102261) and ARC-LIEF (190100080) grants, as well as a research grant from Baidu on autonomous driving.  The authors gratefully acknowledge the GPUs donated by NVIDIA Corporation.  We thank all anonymous reviewers and ACs for their constructive comments.


{\small
\bibliographystyle{ieee_fullname}
\bibliography{egbib}
}

\end{document}